\documentclass[]{article}

\usepackage{tikz}
\usepackage{amsmath}
\usepackage{amsfonts}
\usepackage{hyperref}
\usepackage{multirow}
\usepackage{todonotes}
\usepackage{hyperref}

\usetikzlibrary{arrows}
\usetikzlibrary{fit}
\usetikzlibrary{calc,shapes}
\definecolor{lavander}{cmyk}{0.75,0.28,0,0}
\definecolor{grn}{cmyk}{0.1,0.28,.75,0}
\def\lav{lavander!90}
\def\grn{grn!90}
\tikzstyle{peers}=[draw,circle,black,bottom color=\lav, top color= white, text=black,minimum width=4pt]
\tikzstyle{epeers}=[draw,rectangle,black,bottom color=\grn, top color=white, text=black,minimum width=4pt]

\begin{document}
\title{Reflexive Regular Equivalence for Bipartite Data}
\author{
  Aaron Gerow\footnote{Faculty of Computer Science, Dalhousie University, Halifax, NS, Canada}~\footnote{Corresponding author:\href{mailto:gerow@dal.ca}{gerow@dal.ca}} \and
  Mingyang Zhou\footnote{Department of Computer Science, University of Chicago, Chicago, USA} \and
  Stan Matwin$^*$ \and
  Feng Shi\footnote{University of North Carolina, Chapel Hill, NS, USA} \\
}
\maketitle

\begin{abstract}
Bipartite data is common in data engineering and brings unique challenges, particularly when it comes to clustering tasks that impose on strong structural assumptions. This work presents an unsupervised method for assessing similarity in bipartite data. Similar to some co-clustering methods, the method is based on \textit{regular equivalence} in graphs. The algorithm uses spectral properties of a bipartite adjacency matrix to estimate similarity in both dimensions. The method is reflexive in that similarity in one dimension is used to inform similarity in the other. Reflexive regular equivalence can also use the structure of transitivities -- in a network sense -- the contribution of which is controlled by the algorithm's only free-parameter, $\alpha$. The method is completely unsupervised and can be used to validate assumptions of co-similarity, which are required but often untested, in co-clustering analyses. Three variants of the method with different normalizations are tested on synthetic data. The method is found to be robust to noise and well-suited to asymmetric co-similar structure, making it particularly informative for cluster analysis and recommendation in bipartite data of unknown structure. In experiments, the convergence and speed of the algorithm are found to be stable for different levels of noise. Real-world data from a network of malaria genes are analyzed, where the similarity produced by the reflexive method is shown to out-perform other measures' ability to correctly classify genes.\footnote{A condensed version of this paper will appear in \textit{Proceedings of the 30th Canadian Conference on Artificial Intelligence, Edmonton, Alberta, Canada}.}
\end{abstract}

\section{Introduction}
Co-similarity is the notion that similarity in one dimension is matched by similarity in some other dimension, and it is a pervading assumption in many bipartite data analysis strategies. Unfortunately, co-similarity is often tacitly assumed in exploratory cluster analysis. In particular, co-clustering -- the simultaneous clustering of rows and columns in a matrix -- is increasingly prominent in a range of applications \cite{Tanay}, but assumes that similarity among rows reciprocated by similarity in at least some columns. A canonical example of co-similar structure is in text analysis where similar words appear in similar documents, thus, the word-document co-occurrence matrix can be permuted in a way that exposes co-similarity \cite{Dhillon}. This structure will be evident in diagonal block structure in the resulting similarity matrices, one for rows and one for columns. Bipartite cluster analysis has grown in popularity, in part, due to the ubiquity of two-dimensional data. These include text data, gene expression networks, consumer co-purchasing data and hypergraphs of social affiliation. Each exhibit co-similar structure where similarity in one dimension is matched by similarity in the other: similar words occur in similar documents, similar genes express similar traits, and so on \cite{Jeh}. The work here describes a method of \textit{assessing} co-similarity using regular equivalence in graphs \cite{Leicht} with a reflexive conception similarity that can use nodes (data-points) local transitive structures. Rather than developing statistical tests for co-similarity, the method is designed to be an intuitive and easy-to-interpret assessment.

This assessment can be thought of a pre-condition to a co-clustering analysis: if there is little to no co-similarity structure, co-clustering will be a pointless exercise, and many algorithms will nonetheless produce clusters of poor quality. Assessing co-similarity, unlike co-clustering methods, will produce similarity in one dimension to evince clustering behavior without requiring it of the other dimension. As we will see, this is particularly useful when a non-clustered dimensions informs, but does not reciprocate clustering of the other. We refer to this kind of data as ``asymmetric'' because clusters would be apparent as off-diagonal blocks whereas symmetric data would have blocks only on the diagonal. Assessing co-similarity typically amounts finding an optimal partitioning across data points treated as links from one dimension to another. Current methods suffer three weaknesses: 1) they require binary-valued data (representing extant edges in a bipartite graph), 2) they need an estimate of the number of clusters and 3) they often fail to incorporate the effect of local transitive structures in and across dimensions. The method described here addresses these weaknesses, but differs from co-clustering in some important ways. Whereas as co-clustering finds clusters across both dimensions, our method iterates between dimensions, using each side alternatingly to inform similarity in the other, providing a decoupled solution for each. As such, reflexive regular equivalence is able to quantify the extent to which similarity in one dimension informs similarity in the other -- a tenuous assumption often imposed by co-clustering. The results also show that adjusting weights on local structures of the algorithm with its only free parameter, it is better able to overcome various levels of noise.

\subsection*{Background}
Bipartite data can be thought of as a network in which vertex similarity has two sides. Measuring co-similarity then amounts to calculating vertex similarity in both dimensions. This can be done with bipartite forms of regular equivalence \cite{Borgatti2,Doreian}, which calculate vertex similarity based on the similarity of two vertices neighbors. Formulating unimodal similarity in this way, one can think of co-similarity as regular equivalence in one dimension that is informed by regular equivalence in the other. That is, two nodes in dimension 1 are similar if they neighbor similar nodes in dimension 1 that are connected to the same nodes in dimension 2, themselves similar to the extent they share similar neighbors in their own dimension. Bipartite regular equivalence makes sense of \textit{within-dimension regularity} and \textit{between-dimension structure}. As we will see, however, allowing the inter-dimensional structural aspect to be ``reflexive'', as we define it below, greatly helps assess co-similarity in a range of noisy, asymmetric data.

We conceive of bipartite data as a two-mode network, variously referred to as a bimodal, bipartite or a two-layer network. In single-mode networks, vertex similarity can be measured in many ways. A common approach is (local) structural equivalence. For example, the Jaccard index defines the similarity, $\sigma$, of two nodes, $i$ and $l$ using their shared neighbors ${{|\Gamma_i \cap \Gamma_j|} \over {|\Gamma_i \cup \Gamma_j|}}$ where $\Gamma_v$ denotes the set of neighbors of $v$. Other measures include the cosine of a pair's connectivity patterns, their Pearson correlation or simply their neighborhoods' overlap (see \cite{Zhang} for a review). Recent efforts have sought to account for structure beyond the set of two nodes' immediate neighbors. \textit{Regular equivalence} treats two nodes as similar to the extent their neighbors are themselves similar \cite{Leicht,Blondel}. This definition is recursive and typically implemented as an iterative algorithm beginning with a random initialization. Regular equivalence is common in social network analysis \cite{Gnatyshak} and has been extended to bipartite structures in a block-modeling framework \cite{Borgatti}. This type of co-similarity was first formalized in a social setting by Ronald Breiger in 1974 \cite{Breiger} where he showed that similar people tend to participate in similar groups and that similar groups connect similar people. Using the adjacency structure, $A$, of $people \times groups$, their similarity can be measured as

\begin{eqnarray*}
S_{people}=AA^{\mathsf{T}}\\
S_{groups}=A^{\mathsf{T}}A
\end{eqnarray*}

\noindent
Conceived of this way, social organization can be metricated for use in models of social search 
\cite{Gerow,Shi}. The method presented here extends regular equivalence and Breiger's notion of co-similarity in bipartite data. Crucially, the method also considers structural equivalence by adding contributions from local transitivity -- two nodes are similar if their shared neighbors are themselves similar \cite{everett}.

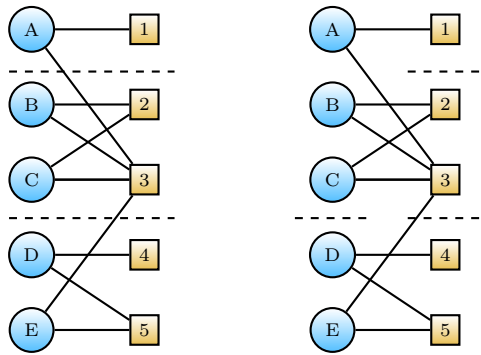
\begin{figure}
	\centering
	  \begin{tikzpicture}[
      shorten >= 0pt,above=9pt,node distance=2.5cm, thick,main node/.style={circle,fill=\lav!20,draw,font=\sffamily\small},
      every newellipse node/.style={inner sep=0pt}
    ]

    \node[peers] (n1) {{\scriptsize A}};
    \node[peers] (n2) [below of=n1,yshift=1.5cm] {{\scriptsize B}};
    \node[peers] (n3) [below of=n2,yshift=1.5cm] {{\scriptsize C}};
    \node[peers] (n4) [below of=n3,yshift=1.5cm] {{\scriptsize D}};
    \node[peers] (n5) [below of=n4,yshift=1.5cm] {{\scriptsize E}};

    \node[epeers] (e1) [right   of=n1,xshift=-1cm] {{\scriptsize 1}};
    \node[epeers] (e2) [below of=e1,yshift=1.5cm]  {{\scriptsize 2}};
    \node[epeers] (e3) [below of=e2,yshift=1.5cm]  {{\scriptsize 3}};
    \node[epeers] (e4) [below of=e3,,yshift=1.5cm] {{\scriptsize 4}};
    \node[epeers] (e5) [below of=e4,yshift=1.5cm]  {{\scriptsize 5}};

    \path[every node/.style={font=\sffamily\small}]
    (n1) edge node {} (e1)
    (n2) edge node {} (e2)
    (n2) edge node {} (e3)
    (n1) edge node {} (e3)
    (n3) edge node {} (e2)
    (n3) edge node {} (e3)
    (n3) edge node {} (e3)
    (n4) edge node {} (e4)
    (n5) edge node {} (e3)
    (n4) edge node {} (e5)
    (n5) edge node {} (e5);

    \draw [dashed] (-.3,-.25) -- (1.9,-.25);
    \draw [dashed] (-.3,-2.2) -- (1.9,-2.2);

    \node[peers] (m1) [right of=e1] {{\scriptsize A}};
    \node[peers] (m2) [below of=m1,yshift=1.5cm] {{\scriptsize B}};
    \node[peers] (m3) [below of=m2,yshift=1.5cm] {{\scriptsize C}};
    \node[peers] (m4) [below of=m3,yshift=1.5cm] {{\scriptsize D}};
    \node[peers] (m5) [below of=m4,yshift=1.5cm] {{\scriptsize E}};

    \node[epeers] (f1) [right of=m1,xshift=-1cm] {{\scriptsize 1}};
    \node[epeers] (f2) [below of=f1,yshift=1.5cm] {{\scriptsize 2}};
    \node[epeers] (f3) [below of=f2,yshift=1.5cm] {{\scriptsize 3}};
    \node[epeers] (f4) [below of=f3,yshift=1.5cm] {{\scriptsize 4}};
    \node[epeers] (f5) [below of=f4,yshift=1.5cm] {{\scriptsize 5}};

    \path[every node/.style={font=\sffamily\small}]
    (m1) edge node {} (f1)
    (m2) edge node {} (f2)
    (m2) edge node {} (f3)
    (m1) edge node {} (f3)
    (m3) edge node {} (f2)
    (m3) edge node {} (f3)
    (m3) edge node {} (f3)
    (m4) edge node {} (f4)
    (m5) edge node {} (f3)
    (m4) edge node {} (f5)
    (m5) edge node {} (f5);

    \draw [dashed] (3.5,-2.2) -- (4.5,-2.2);
    \draw [dashed] (5,-.25) -- (6,-.25);
    \draw [dashed] (5,-2.2) -- (6,-2.2);
  \end{tikzpicture}
  \caption{The result of the same example bipartite network submitted to a typical co-clustering algorithm (left; a), where the result is a set of partitions (dotted lines) that group vertices across both modes. On the right (b) is a hypothetical grouping based on similarities produced by reflexive regular equivalence. In the reflexive situation, there may not be a one-to-one mapping between the clusters across modes and the partitions are exposed by similarity, not computed by the algorithm.}
  \label{fig:fig1}
\end{figure}

Measuring similarity in bipartite data is related to co-clustering tasks: both produce pairwise equivalences for nodes in both modes that can expose potential clustering. Our goal here is not to find partitions that optimally group nodes across dimensions, but instead, to measure similarity in either mode of the data using information from the other (and only by extension, any clustered structure in the data). Figure \ref{fig:fig1} depicts a bipartite network and hypothetical co-clustering solution (a) and a potential solution based on reflexive regular equivalence (b). In co-clustering, partitions are computed by assessing the clusters themselves, whereas in the our method, clusters are \textit{potential} groupings based on similarity. Whereas most co-clustering algorithms produce a solution for a specified number of clusters, of potentially poor quality, the reflexive case simply produces similarity matrices for each mode of the graph.

Although they are different from the method proposed here, co-clustering algorithms -- often referred to as ``biclustering'' -- have much in common with reflexive regular equivalence. As mentioned, co-clustering algorithms yield partitions across the dimensions of bipartite data that group self-similar rows with self-similar columns. Co-clustering assumes co-similarity just like clustering assumes groups of similar nodes (ie. data-points). The method we propose produced decoupled partitions across bipartite data by \textit{using} similarity information across and between dimensions. \textsc{samba} is a co-clustering technique which uses subgraph detection \cite{Tanay2} and incorporates structure on both sides of bipartite data, such as gene-function graphs. However, \textsc{samba} only uses local structure without regular equivalence: it does not implement the assumption that shared neighbors of similar nodes should themselves be similar. Another co-clustering method uses a graph-drawing scheme that relies on bipartite structural properties to minimize the number of inter-mode edges that cross one another \cite{Abdullah}. Again, such structural techniques rely primarily on first-order structure from both modes of the bipartite graph. More recent advances in co-clustering research have offered agglomerative methods \cite{Codocedo,Pensa,Pio}, producing hierarchically organized clusters, or on accommodating clusters with fuzzy membership \cite{DeFranca,Pio,Yan} and models that learn co-clusters under supervision \cite{Meng,Teng}. These advances help address some weaknesses in the basic co-clustering strategy, but they continue to rely on the assumption of co-similarity. The method we propose provides a way to assess co-similarity in each mode of bipartite data. In some cases, the results will justify a co-clustering strategy (when similarity in one dimension is tightly coupled to the other) and in other cases it will yield similarity in one dimension that is \textit{informed by not coupled to} similarity in the other.

\section{Method}
Our reflexive equivalence method is an unsupervised approach to measuring vertex similarity in a bipartite network. This notion is equivalent to a pairwise similarity metric that operates on row- and column-vectors of the adjacency structure in bimodal data. The network depicted in Figure \ref{fig:fig1} can be represented as an adjacency matrix:

{\small
  \begin{center}
    \begin{tabular}{c|c c c c c|}
      \multicolumn{1}{r}{} &
      \multicolumn{1}{r}{\textbf{1}} &
      \multicolumn{1}{r}{\textbf{2}} &
      \multicolumn{1}{r}{\textbf{3}} &
      \multicolumn{1}{r}{\textbf{4}} &
      \multicolumn{1}{r}{\textbf{5}} \\
      \cline{2-6}
      \textbf{A} & * & - & * & - & - \\
      \textbf{B} & - & * & * & - & - \\
      \textbf{C} & * & - & * & - & - \\
      \textbf{D} & - & - & - & * & * \\
      \textbf{E} & - & - & * & - & * \\
      \cline{2-6}
    \end{tabular}
  \end{center}
}

\noindent
where a * represents an extant edge between nodes. In this way, bipartite networks are equivalent to what are normally treated simply as rows and columns in a matrix. Our method iterates between each mode of the data incorporating information from the other. Note there is no restriction to binary-valued data in the adjacency structure: edges in the network may be weighted.

\subsection{Reflexive Regular Equivalence}
Let $G$ be a bipartite graph with two sets of nodes, $V$ and $V'$. Let $A$ be the adjacency matrix of dimension $|V|$ by $|V'|$, where $A_{ij}>0$ represents how strongly $i\in V$ is connected to $j \in V'$. Let $S$ be a $|V|$ by $|V|$ matrix where entry $S_{ij}$ denotes the similarity between vertices $i$ and $j$ in $V$, and $S'$ be a $|V'|$ by $|V'|$ matrix where entry $S'_{ij}$ denotes the similarity between $i$ and $j$ in $V'$.  Assuming similarity between two nodes $i$ and $j$ in one type is determined by the similarity between any neighbor of $i$ and any neighbor of $j$ in the other type, then

\begin{eqnarray}
S_{ij}=\sum_k\sum_l A_{ik}A_{jl}S'_{kl} \label{eq:s_ij}, \\
S'_{ij}=\sum_k\sum_l A_{ki}A_{lj}S_{kl} \label{eq:sp_ij},
\end{eqnarray}

\noindent
or in matrix form:

\begin{align} 
S &=AS'A^{\mathsf{T}} \label{eq:s}, \\
S'&=A^{\mathsf{T}}SA. \label{eq:sp}
\end{align} 

This provides an iterative procedure with which to infer similarity between nodes in the adjacency matrix. Starting from a random $S$ or $S'$ matrix, one can apply equations \ref{eq:s} and \ref{eq:sp} iteratively until convergence of $\|S\|_{\mathsf{F}}$ and $\|S'\|_{\mathsf{F}}$. This formulation is similar to spectral clustering in a unipartite setting \cite{Kluger,Breiger}. 

\subsection{Incorporating Transitivity}
Equations \ref{eq:s_ij} and \ref{eq:sp_ij} treat all neighbors equally, regardless of their structural importance to the pair of nodes under investigation. Hence, we weight similarity between common neighbors more than non-common neighbors. Denoting neighbors of $i$ and $j$ by $\Gamma_i$ and $\Gamma_j$ respectively, we define this similarity as follows:

{\small
\begin{eqnarray}
S_{ij}=(1-\alpha) [\sum_{k\in(\Gamma_i-\Gamma_i\cap \Gamma_j)}\sum_{l\in \Gamma_j}S'_{kl} + \sum_{k\in \Gamma_i\cap \Gamma_j}\sum_{l\in(\Gamma_j-\Gamma_i\cap \Gamma_j)}S'_{kl}] \notag \\ + \sum_{k\in \Gamma_i\cap \Gamma_j}\sum_{l\in \Gamma_i\cap \Gamma_j}S'_{kl},\\
S'_{ij}=(1-\alpha) [\sum_{k\in(\Gamma_i-\Gamma_i\cap \Gamma_j)}\sum_{l\in \Gamma_j}S_{kl} + \sum_{k\in \Gamma_i\cap \Gamma_j}\sum_{l\in(\Gamma_j-\Gamma_i\cap \Gamma_j)}S_{kl}] \notag \\ + \sum_{k\in \Gamma_i\cap \Gamma_j}\sum_{l\in \Gamma_i\cap \Gamma_j}S_{kl}.
\end{eqnarray}
}

\noindent
This combines notions of structural equivalence, regular equivalence and reflexivity into a single model. We add a parameter $\alpha$ to balance the contributions from non-common and common neighbors. More precisely, the contribution from non-common neighbors is discounted by a factor of $(1-\alpha)$. Rearranging the terms, the equations can be rewritten as

{\small
\begin{eqnarray}
S_{ij}=(1-\alpha)\sum_k\sum_l A_{ik}A_{jl}S'_{kl}+\alpha\sum_k\sum_l A_{ik}A_{jk}A_{il}A_{jl}S'_{kl} \label{eq:s_ij_com},\\
S'_{ij}=(1-\alpha)\sum_k\sum_l A_{ki}A_{lj}S_{kl}+\alpha\sum_k\sum_l A_{ki}A_{kj}A_{li}A_{lj}S_{kl} \label{eq:sp_ij_com},
\end{eqnarray}
}

\noindent
or in matrix form:

\begin{align}
S &= (1-\alpha) AS'A^\mathsf{T}+ \alpha (A \otimes A^\mathsf{T}) \cdot S' \cdot(A \otimes A^\mathsf{T})^\mathsf{T} \label{eq:s_com}, \\
S' &=(1-\alpha) A^\mathsf{T}SA+  \alpha (A^\mathsf{T}\otimes A) \cdot S \cdot (A^\mathsf{T}\otimes A)^\mathsf{T}   \label{eq:sp_com}, 
\end{align}

\noindent
where $B=A\otimes A^\mathsf{T}$ is a third-order tensor with $B_{ijk}=A_{ik} A^\mathsf{T}_{kj}$, and ($\cdot$) denotes inner product along the corresponding dimension (as conventionally defined in tensor algebra). Regardless of $\alpha$, the coupling between Eqs. \ref{eq:s_com} and \ref{eq:sp_com} exemplifies reflexivity between the two dimensions. Additionally, the effect from local structure is controlled by $\alpha$. When $\alpha=0$, the method is a bipartite form of typical regular equivalence; as  $\alpha$ increases, similarity between common neighbors (and hence local transitivity)  plays a larger role. 

\subsection{Implementation}
Two points are worth noting in the implementation of the method above. First, as a pre-processing step, entries of  $A$ are normalized to lie between 0 and 1 by dividing each row by its sum. In addition to helping ensure convergence in a numerical sense, this normalization has a physical analogue. For example, stop words -- semantically irrelevant words -- appear in documents, and would have many neighbors in the word-document adjacency structure. If the similarity between two documents were calculated based on the similarities between words in the two documents, then stop words carry no meaningful information and their contribution should be discounted proportional to their connectivity (their degree, in a network sense). Normalizing $A$ diminishes the contribution of structurally irrelevant data. To help prevent numerical errors, $S$ and $S'$ are also normalized at every iteration. This also ensures that the results are interpretable on the same scale as $A$. In each iteration, after computing the similarity matrix (which is symmetric) its values are normalized by its $L_1$-, $L_2$- or $L_{\infty}$-vector norm. Lastly, $S$ and $S'$ are initialized randomly, a strategy that works well in single-mode regular equivalence \cite{Leicht}.

As we will see, similarity structure in either dimension of $A$ will produce block-diagonal structure in the resulting $S$ or $S'$. This leads us to our first evaluation where $A$ has known structure (block-diagonal) comparing our results to pairwise metrics that operates on the rows and columns of $A$.

\section{Results}

\subsection{Synthetic Data}
We begin by evaluating our method on data with known structure. A set of semi-random $n \times m$ versions of $A$ were generated with diagonal blocks of varying sizes. Computing reflexive similarity for $A$ will produce $S$ ($n \times n$) and $S'$ ($m \times m$) which should have diagonal blocks of equal size and number as $A$ and proportional to their row- and column-wise sizes. This is a trivial case because we know the results from spectral clustering (when $\alpha=0$). Instead, the test is to run the algorithm on randomly permuted versions of the adjacency matrix, $\hat{A}$, after which if we apply the original ordering of $A$ to $\hat{S}$ and $\hat{S'}$, the results should be similar (see Figure 2). We assess the difference in results in a stringent way using their Euclidean distance:

\begin{equation}
\mu =  {{1}\over{2}} \|S - \hat{S}\|_{\mathsf{F}} {{1} \over {|S|}} +  {{1}\over{2}} \|S' - \hat{S'}\|_{\mathsf{F}} {{1} \over {|S'|}} . \label{eq:performance}
\end{equation}

\noindent
Here, $\mu$ is simply the mean difference between $\hat{S}$ and $\hat{S}'$, generated from our method on $\hat{A}$, and the ``ground truth'' $S$ and $S'$ from $A$. In the perfect case, $\mu$ will be 0 because the results in both dimensions of $\hat{A}$ are the same as on $A$.

\begin{figure}
  \centering
  \includegraphics[scale=0.3]{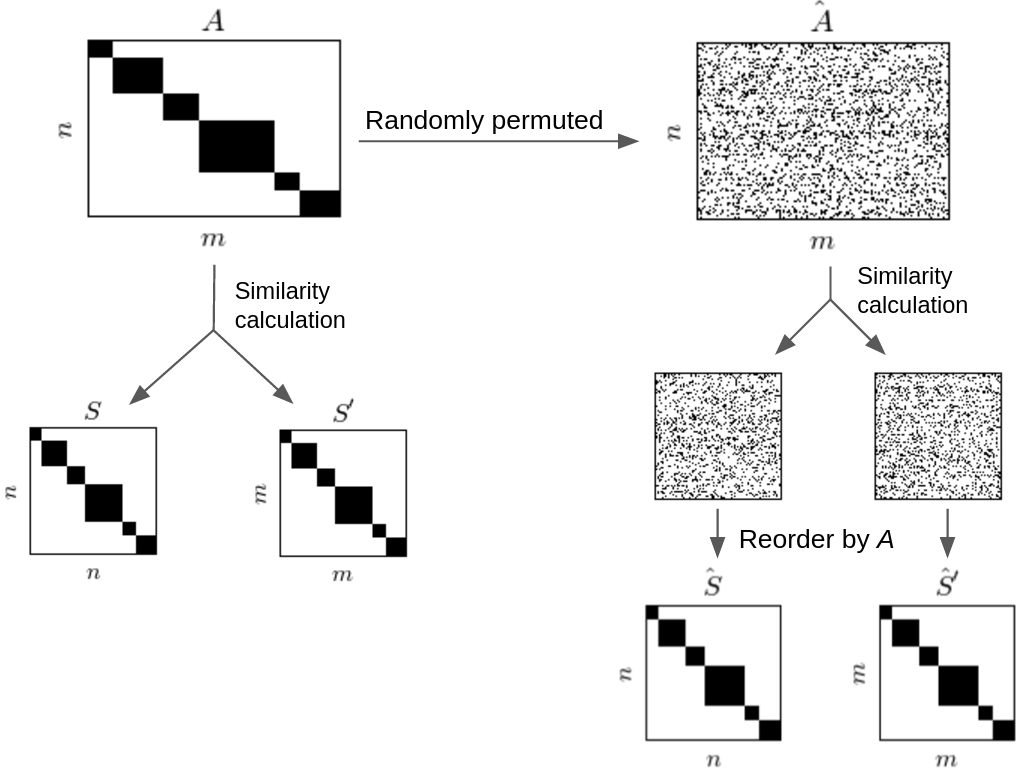}
  \label{fig:eval1}
  \caption{Computing similarity on the permuted adjacency matrix, $\hat{A}$ should produce results similar to those computed for $A$.}
\end{figure}

\begin{figure}
  \centering
  \includegraphics[scale=0.75]{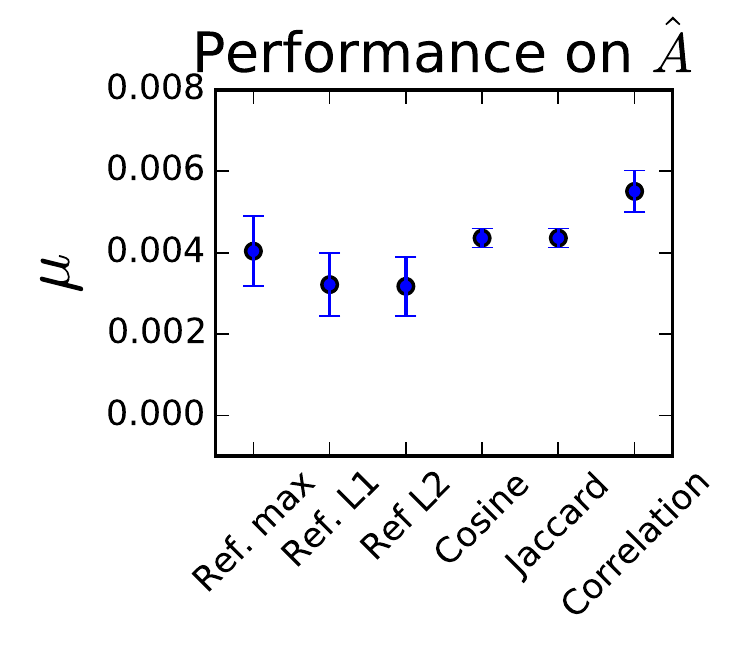}
  \caption{Performance ($\mu$; lower is better) on $\hat{A}$ using the reflexive similarity method with different normalizations and pairwise metrics that operate on the rows and columns of $A$. Correlation refers to the Pearson correlation coefficient. Error-bars are $\pm2$ s.e. of the mean across ten versions of $A$, each built to have block-diagonal structure. In the reflexive methods, values of $\alpha$ did not change the results.}
  \label{fig:Ahat_perf}
\end{figure}

We compare reflexive similarity to three pairwise metrics calculated using the rows and columns of $A$. The results show that the reflexive methods perform best, but not beyond $\pm2$ s.e. of the other methods (Figure \ref{fig:Ahat_perf}). The other methods have no free parameters and for the three variants of the reflexive method, we found that $\alpha$ had no effect on the results. This behavior is expected because all node interactions are purely local (cf., the block diagonal structure), and hence more contribution from local structure (increasing $\alpha$) does not impact the results. In the subsequent experiments, we explore the effects of noise (which will break symmetry in $A$) and situations where where $A$ has unbalanced block structure.

\subsubsection*{Robustness to Noise}
Real data is often noisy, and it is important to assess performance in situations with less discernible structures. To do this, we add noise to $A$ of the form $\mathcal{N}(0,\sigma)$ to produce $\tilde{A}$, which is submitted to the same test as above. Results are reported in terms of $\mu$ (Eq. \ref{eq:performance}) for variants of the reflexive method and pairwise metrics. Note that adding noise breaks the element-wise symmetry of $A$, but the underlying ``true'' block-structure remains symmetric (on diagonal).

\begin{figure}
  \centering
  \includegraphics[scale=0.75]{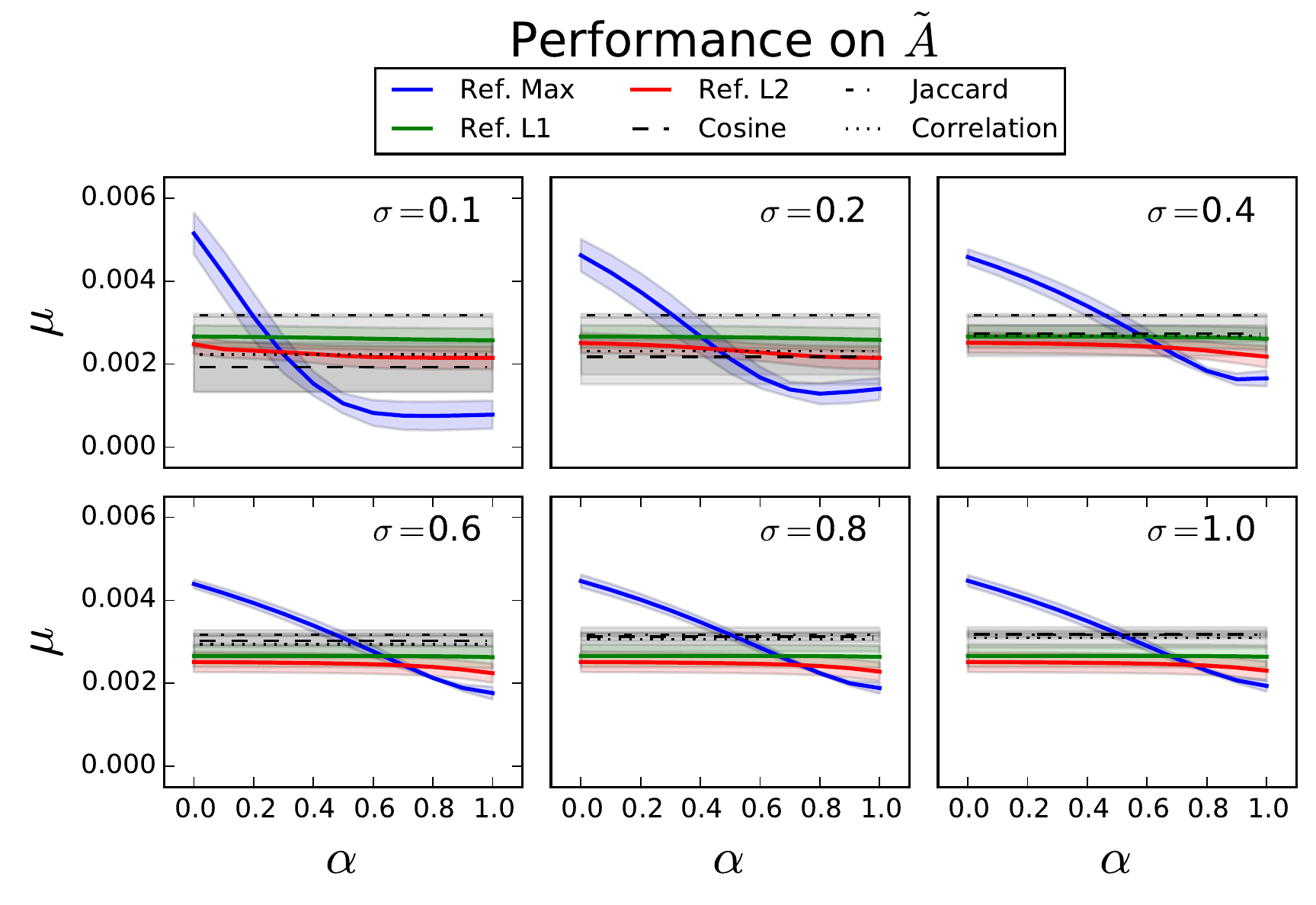}
  \caption{Performance ($\mu$; Eq. \ref{eq:performance}; lower is better) with respect to the level of noise, $\sigma$ in $\tilde{A}$ for different values of $\alpha$. Error-bands are $\pm2$ s.e. of the mean over ten versions of $A$.}
  \label{fig:perf2}
\end{figure}

Figure \ref{fig:perf2} shows results with respect to noise, variant and $\alpha$. With little noise, the $L_{\infty}$-norm variant of reflexive similarity outperforms other methods when $\alpha > 0.5$. As the level of noise in $\tilde{A}$ is increased, the pairwise metrics perform less and less well, as do the reflexive methods with $\alpha=0$. This confirms that the reflexive method is well-suited to finding similarity structure in both dimensions of noisy data. The results also show that the $L_{\infty}$-norm variant, with $\alpha=1$ is the best overall for noisy, symmetric structures. But perhaps most importantly, the fact that increased $\alpha$ leads to better performance shows that exploiting local and global information in noisy data provides a significant boost in performance over metrics that only use similarity in a single dimension.

\subsubsection*{Unbalanced Co-similarity}
\label{sec:unbalanced}
\begin{figure}
  \centering
  \includegraphics[scale=0.4]{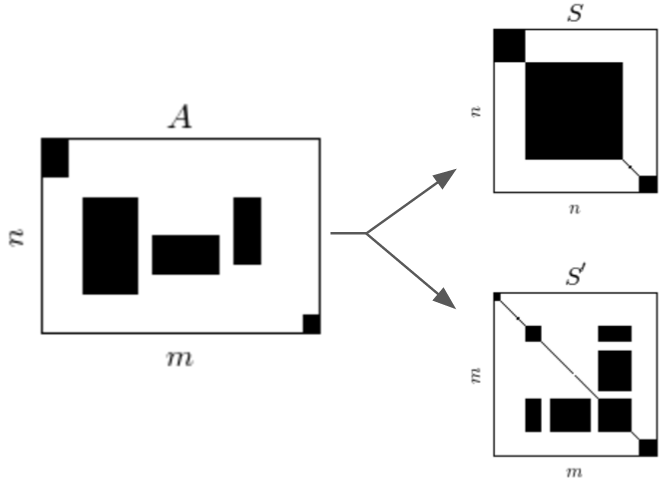}
  \caption{Evaluating bipartite data with unbalanced co-similarity structure, the resulting similarity matrices will expose different clusters corresponding to dimensions in $A$. The off-diagonal blocks in $S'$ correspond to the off-diagonal blocks in the $m$-dimension of $A$, one of which lies party on the diagonal.}
  \label{fig:eval2}
\end{figure}

\begin{figure}
  \centering
  \includegraphics[scale=0.6]{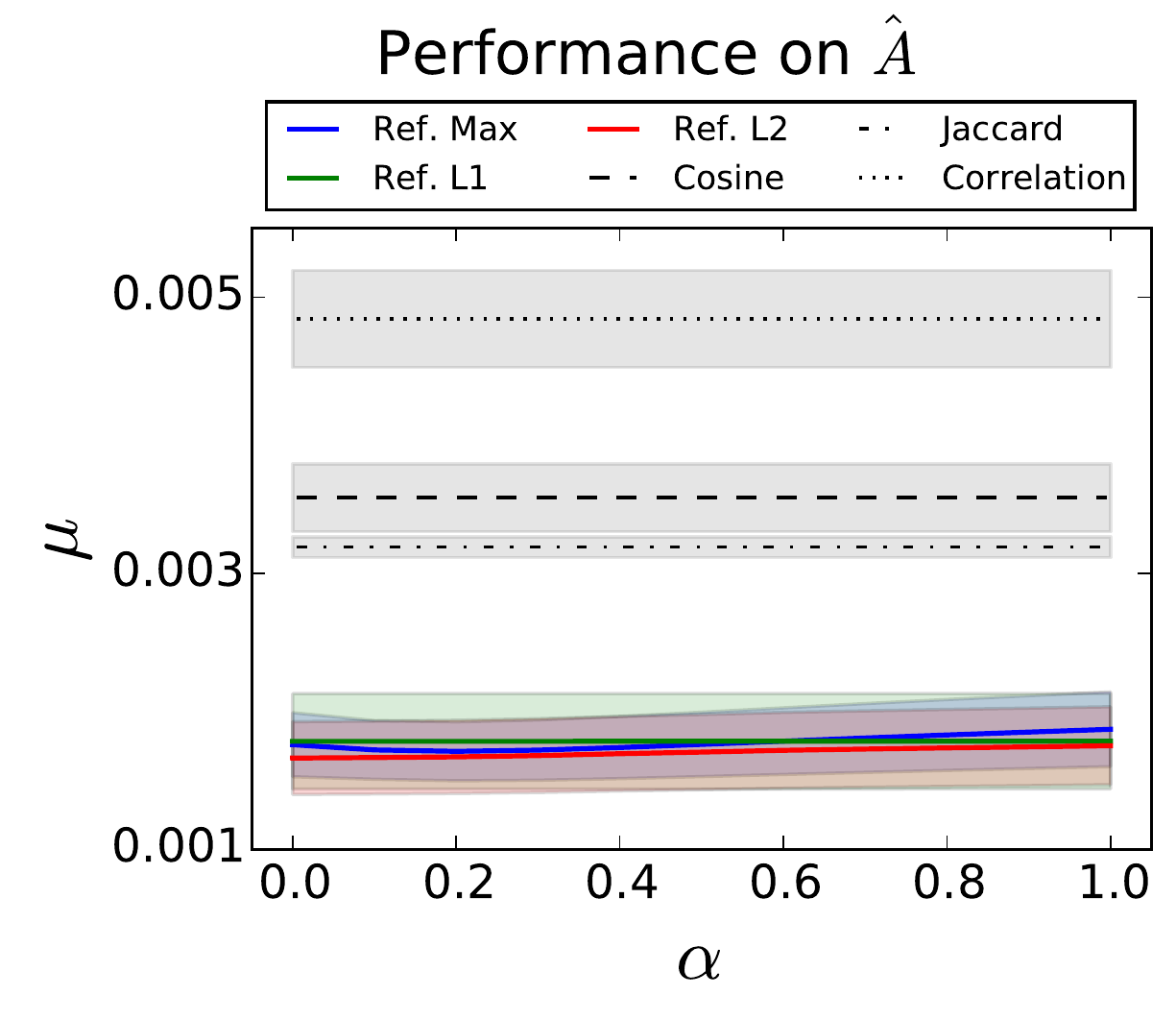}
  \caption{Performance ($\mu$; lower is better) on versions of $\hat{A}$ with unbalanced structure. Error-bands are $\pm2$ s.e. of mean across ten version of $A$.}
  \label{fig:b_hat-perf}
\end{figure}

Because any clusters exposed by similarity in $S$ and $S'$ are not coupled across rows and columns in $A$, as they are in co-clustering tasks, it is possible for reflexive similarity to find unbalanced co-similarity structure. For example, Figure \ref{fig:eval2} depicts data with three row-wise blocks in $A$, but five across its columns. With this kind of data, the reflexive method will produce similarity matrices with different structure. For this task, another set of semi-random versions of $A$ were generated with unbalanced block structure, which were then randomly permuted to get $\hat{A}$. Results on $\hat{A}$ were compared to results from $A$ as was done in the first evaluation. Figure \ref{fig:b_hat-perf} shows the results on the three variants of reflexive regular equivalence and the pairwise metrics. The results show that variants of reflexive similarity outperform the pairwise metrics. Between the different normalizations, there is no significant difference in performance. Likewise, different values of $\alpha$ for the $L_{\infty}$-norm variant show only slight variation.

\begin{figure}
  \centering
  \includegraphics[scale=0.7]{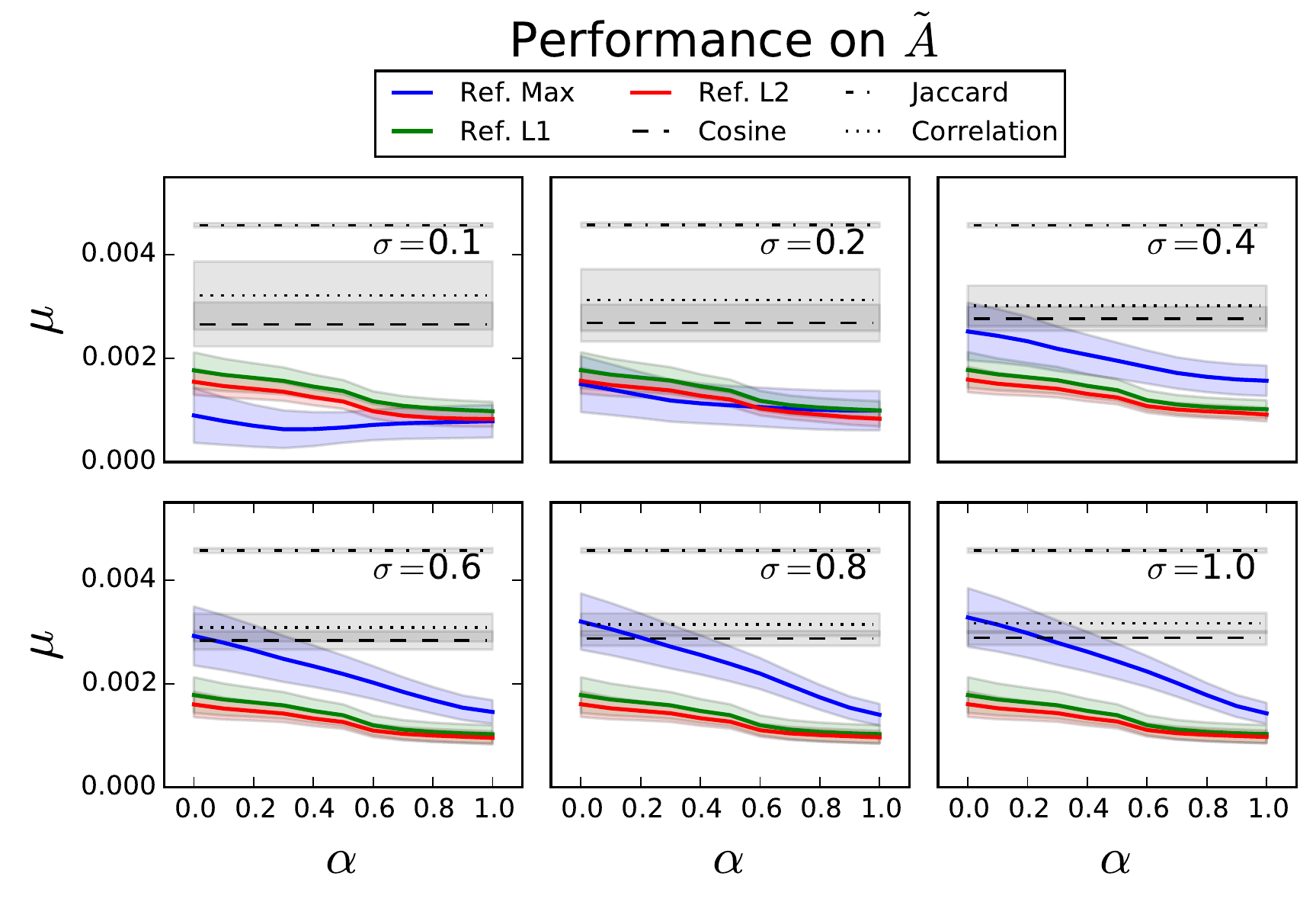}
  \caption{Performance on $\tilde{A}$ with asymmetric co-similarity structure. Error-bands are $\pm2$ s.e. of the mean over ten versions of $A$.}
  \label{fig:b_tilde-perf}
\end{figure}

Overcoming noise in $A$ is a crucial aspect to our method's performance, and it was found that $\alpha$ helped the model in this regard when $A$ is symmetric. The same task was run for asymmetric versions of $A$ where noise was added proportional to $\mathcal{N}(0,\sigma)$ to yield $\tilde{A}$. Figure \ref{fig:b_tilde-perf} shows performance for different intensities of noise and values of $\alpha$. The results show that even after adding a small amount of noise to the adjacency structure, the reflexive variants significantly outperform the pairwise metrics. In particular, for low levels of noise, the $L_{\infty}$-norm variant performs best, but for higher levels of noise ($\sigma > 0.2$) the $L_1$- and $L_2$-norms are better. In all variants, higher values of $\alpha$ yield better performance. Overall, this implies that local structure is a useful feature even when co-similarity structure is not symmetric. These evaluations also show that on synthetic data, where the structure of $A$ is known ahead of time, that reflexive similarity is indeed able to leverage inter-dimensional similarity in noisy and asymmetric data to provide better results than methods restricted to one dimension. After an experimental assessment of the algorithm's convergence behavior and runtime, we turn to a real-world example of bipartite similarity analysis in genetic data.

\subsection{Convergence Behavior \& Runtime}
\begin{figure}
  \centering
  \includegraphics[scale=0.65]{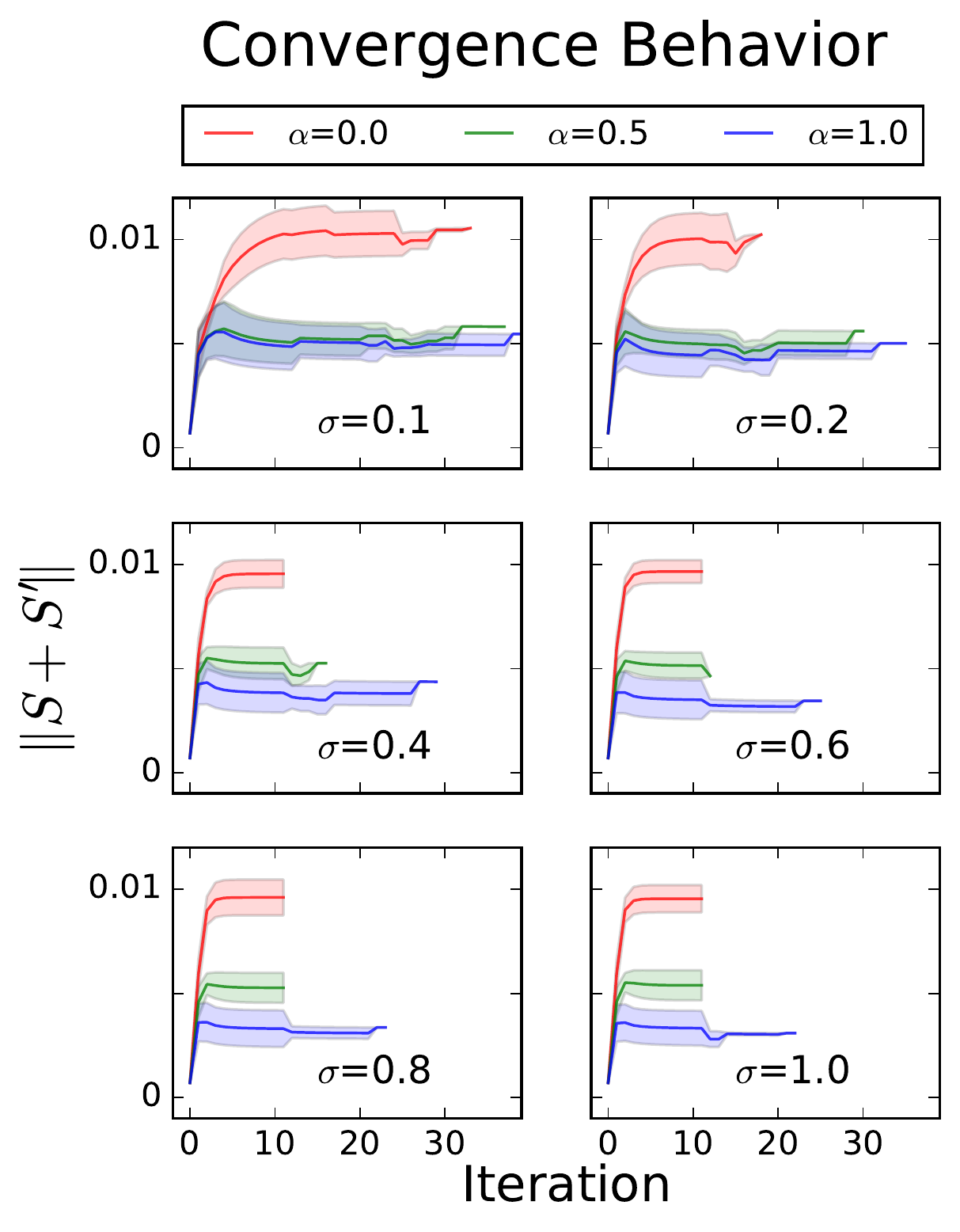}
  \caption{Convergence behavior of the $L_{\infty}$ variant of reflexive similarity for $\alpha \in \{0,.5,1\}$ and levels of noise ($\mathcal{N}(0,\sigma)$ for $\sigma \in \{.1,.2,.4,.6,.8,1\}$) on ten symmetric versions of $A$.}
  \label{fig:convergence_sym}
\end{figure}

\begin{figure}
  \centering
  \includegraphics[scale=0.65]{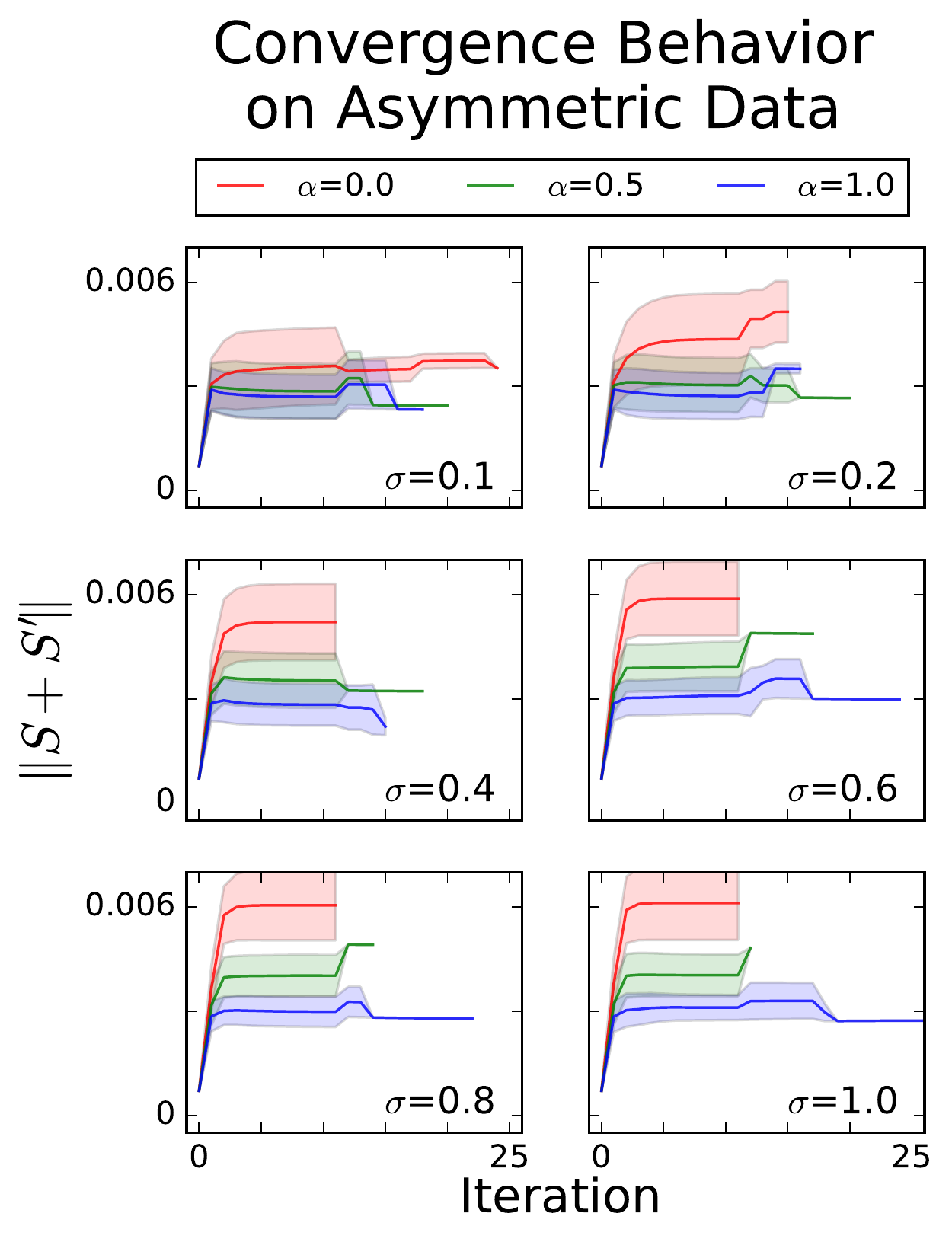}
  \caption{Convergence behavior with different values of $\alpha \in \{0,.5,1\}$ and levels of noise ($\sigma \in \{.1,.2,.4,.6,.8,1\}$) on ten asymmetric versions of $A$.}
  \label{fig:convergence_asym}
  \end{figure}

Convergence of the norm for $S$ and $S'$ is not guaranteed in any fixed number of iterations. Here, we look at the rate and shape of convergence in these norms in an experimental setting. The same data described in the previous section were used to assess convergence in $\|S\|_{\mathsf{F}}$ and $\|S'\|_{\mathsf{F}}$ by randomly permuting $A$ into $\hat{A}$ and running the algorithm. In all cases the convergence threshold was set to 1\textsc{e}-5. Figure \ref{fig:convergence_sym} plots $\|S+S'\|_{\mathsf{F}}$ over each full iteration of the algorithm for various values of $\alpha$ and with various intensities of noise added to $A$. The method requires more iterations to converge for higher values of $\alpha$ as more information from local transitivities is used. A pattern related to sparsity can also be observed: as the noise is increased (greater values of $\sigma$) sparsity in $\hat{A}$ decreases, resulting in fewer iterations before convergence. This is particularly interesting because it shows that noisy data will not necessarily require more iterations to converge. Figure \ref{fig:convergence_asym} plots the results on the asymmetric versions of $A$ described above. With asymmetric data, a similar pattern is observed with respect to noise that reduces sparsity, but we also see that, in general, more iterations are required when $A$ is asymmetric.

\begin{table}
  \centering
  \begin{tabular}{r c r r r r}
    \hline
    \multicolumn{6}{c}{\textbf{Symmetric} $\mathbf{A}$} \\
    $\mathbf{\alpha}$ & \textbf{Noise (}$\mathbf{\sigma}$\textbf{)} & \textbf{Iterations} & \textbf{Time per Iter.} & \textbf{Total} & \textbf{s.e. Total} \\
    \hline
    \multirow{6}{*}{0}
    & 0.0 & 70.9 & 0.054 & 3.82 & 1.03 \\
    & 0.2 & 15.4 & 0.220 & 3.38 & 0.55 \\
    & 0.4 & 12.0 & 0.214 & 2.57 & 0.32 \\
    & 0.6 & 12.0 & 0.214 & 2.57 & 0.33 \\
    & 0.8 & 12.0 & 0.214 & 2.57 & 0.32 \\
    & 1.0 & 12.0 & 0.216 & 2.59 & 0.32 \\
    \hline
    \multirow{6}{*}{.5}
    & 0.0 & 70.9 & 0.091 & 6.44 & 1.91 \\
    & 0.2 & 19.1 & 0.293 & 5.60 & 0.75 \\
    & 0.4 & 13.1 & 0.299 & 3.91 & 0.52 \\
    & 0.6 & 12.3 & 0.300 & 3.69 & 0.53 \\
    & 0.8 & 12.1 & 0.309 & 3.73 & 0.57 \\
    & 1.0 & 12.0 & 0.300 & 3.61 & 0.53 \\
    \hline
    \multirow{6}{*}{1}
    & 0.0 & 70.9 & 0.088 & 6.25 & 1.84 \\
    & 0.2 & 20.6 & 0.280 & 5.76 & 0.78 \\
    & 0.4 & 15.8 & 0.277 & 4.37 & 0.59 \\
    & 0.6 & 14.8 & 0.283 & 4.19 & 0.59 \\
    & 0.8 & 14.4 & 0.286 & 4.11 & 0.55 \\
    & 1.0 & 14.3 & 0.285 & 4.07 & 0.52 \\
    \hline
    \\
    \multicolumn{6}{c}{\textbf{Asymmetric} $\mathbf{A}$} \\
    $\mathbf{\alpha}$ & \textbf{Noise (}$\mathbf{\sigma}$\textbf{)} & \textbf{Iterations} & \textbf{Time per Iter.} & \textbf{Total} & \textbf{s.e. Total} \\
    \hline
    \multirow{6}{*}{0}
    & 0.0 & 14.8 & 0.132 & 1.95 & 0.31 \\
    & 0.2 & 13.2 & 0.318 & 4.20 & 0.26 \\
    & 0.4 & 12.0 & 0.315 & 3.78 & 0.18 \\
    & 0.6 & 12.0 & 0.319 & 3.82 & 0.20 \\
    & 0.8 & 12.0 & 0.315 & 3.78 & 0.20 \\
    & 1.0 & 12.0 & 0.315 & 3.78 & 0.19 \\
    \hline
    \multirow{6}{*}{.5}
    & 0.0 & 18.8 & 0.212 & 3.98 & 1.41 \\
    & 0.2 & 14.8 & 0.489 & 7.23 & 1.01 \\
    & 0.4 & 12.7 & 0.493 & 6.26 & 0.55 \\
    & 0.6 & 12.6 & 0.519 & 6.54 & 0.44 \\
    & 0.8 & 12.2 & 0.515 & 6.29 & 0.42 \\
    & 1.0 & 12.1 & 0.552 & 6.67 & 0.46 \\
    \hline
    \multirow{6}{*}{1}
    & 0.0 & 15.1 & 0.188 & 2.84 & 0.49 \\
    & 0.2 & 13.6 & 0.518 & 7.05 & 0.82 \\
    & 0.4 & 14.5 & 0.491 & 7.12 & 0.75 \\
    & 0.6 & 13.9 & 0.487 & 6.77 & 0.57 \\
    & 0.8 & 13.7 & 0.511 & 7.00 & 0.88 \\
    & 1.0 & 13.2 & 0.503 & 6.63 & 0.75 \\
    \hline
    \\
    \end{tabular}
  \caption{Runtime for symmetric and asymmetric data with different amounts of added noise. Values are the average of ten versions of $A$.}
  \label{tab:runtime}
\end{table}

Each iteration is the result of a series of matrix operations, and each iteration may take a different amount of time depending on the data. Table \ref{tab:runtime} lists the runtime in various configurations on symmetric and unbalanced versions of $A$. In the balanced case with $\alpha$=0, despite taking fewer iterations on noisy data, each individual iteration takes longer leading to a modest increase on the denser, noisy data. As inter-dimensional data is incorporated with greater values of $\alpha$, the required number of iterations increases slightly for noisy data and they take slightly longer individually. On the asymmetric data, iterations tend to take longer, again as noise and $\alpha$ increase. The runtime behavior of the reflexive similarity method is fairly expected: incorporating inter-dimensional information and noise both require more numerical operations for each matrix operation, but density in $A$ reduces the number of iterations before the similarity matrices stabilize.

\begin{figure}
  \centering
  \includegraphics[scale=0.5]{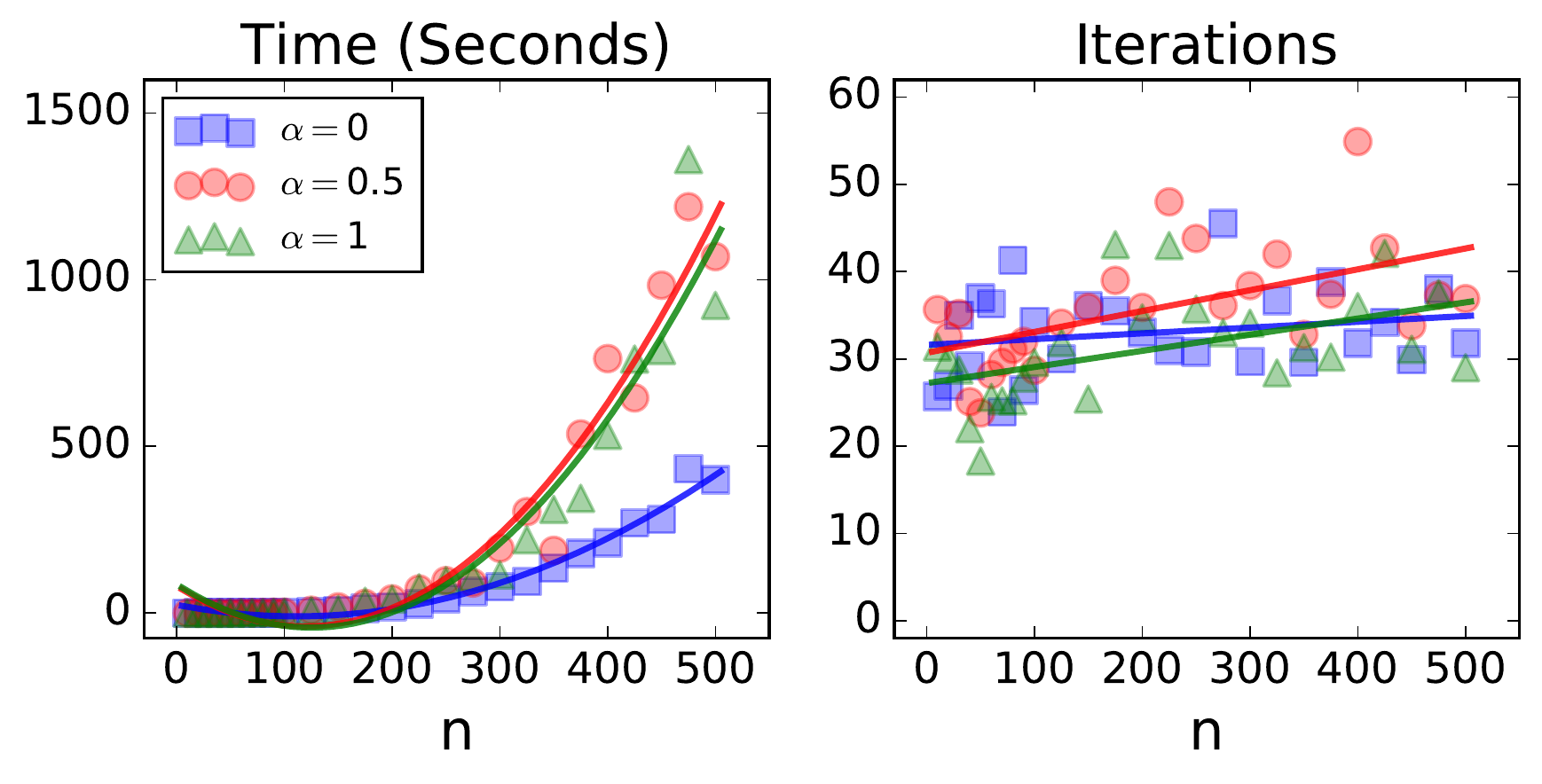}
  \caption{The algorithm's scaling behavior with respect to the size of $A$ in terms of iterations to convergence and the time per iteration. On the left is seconds per iteration and 2nd-order polynomial best-fit lines for three values of $\alpha$. On the right is the number of iterations before convergence with linear best-fits. In all cases, points are the average of ten random versions of $A$ of the size $n \times n$ with random diagonal block-structure.}
  \label{fig:size_scaling}
\end{figure}

The algorithm scales predictably with respect to the dimensions of $A$. Figure \ref{fig:size_scaling} shows the algorithm's scaling behavior with respect to iteration and time for $n \times n$ adjacency structures for $n$ from 10 to 500. Given the element-wise operations in Eqs. 9 and 10, we can expect geometric scaling with respect to total runtime. Indeed, the number of iterations scales linearly at approximately two iterations per $\Delta n=100$ (Figure \ref{fig:size_scaling}), but the runtime scales polynomially. Increased values of $\alpha$ do not significantly effect the number of iterations before convergence, but they do change the total runtime.

\subsection{Analysis of Malaria \textit{Var} Genes}
The evaluations above were performed on synthetic data, designed to simulate various kinds of data and to assess the convergence behavior of the method. Because the data were generated semi-randomly and because the evaluation criteria is based on a the simple \textsc{f}-norm of the similarity matrices, it may not be a fair representation of real-world situations. To more clearly demonstrate the usefulness of the proposed technique, we show that it is useful in practice. Here, we test reflexive similarity on empirical data from malaria \textit{var} genes\footnote{Available at \url{danlarremore.com/bipartiteSBM}.}.

\begin{figure}
  \centering
  \includegraphics[scale=0.25]{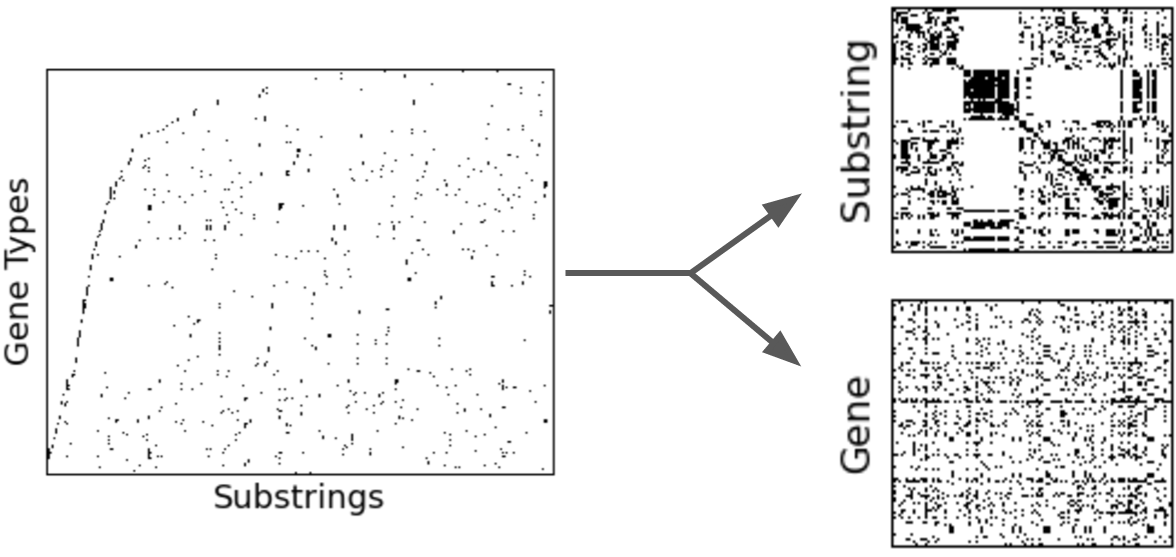}
  \caption{The malaria \textit{var} gene data is a sparse, bipartite matrix of $gene \times substring$ (left), which is processed the same way the synthetic data was, producing similarity matrices in both dimensions. The quality of $S_{gene}$ can be assessed using the HVR classifications offered by \cite{Larremore} ($S_{substring}$ is not analyzed). Note that there may be a permutation of $S_{gene}$ that reveals block structure -- something a co-clustering algorithm might expose -- but it is irrelevant here because the values alone should evince the HVR class structure.}
  \label{fig:malaria_eval}
\end{figure}

Clustering and co-clustering analyses have become an important part of large-scale genetic analysis. In particular, co-clustering the character strings that define amino acids with their gene types, expressions and biological network data has provided valuable insights in systems biology \cite{Hanisch,Pio}. In this analysis, the data codes genes and the amino acid substrings that designate them, but which vary slightly across genes \cite{Larremore,Thomas}. The adjacency matrix consists of bipartite data representing 297 genes and 803 substrings, which is known to exhibit co-similarity. The matrix is binary-valued and relatively sparse (Figure \ref{fig:malaria_eval}; 1.4\% non-zeros). This structure can be thought of as a bipartite network -- not unlike Figure \ref{fig:fig1} -- where genes are connected to their identifying strings. This structure is known to be co-similar and, crucially, it comes with an empirical ground-truth: \textit{highly variable regions} (HVRs) into which each gene is classified on a biological basis. This will allow the similarity produced on the gene-substring adjacency matrix to be qualified by how well the resulting similarity tends to correctly pair genes into the same HVR.

\begin{figure}
  \centering
  \includegraphics[scale=0.55]{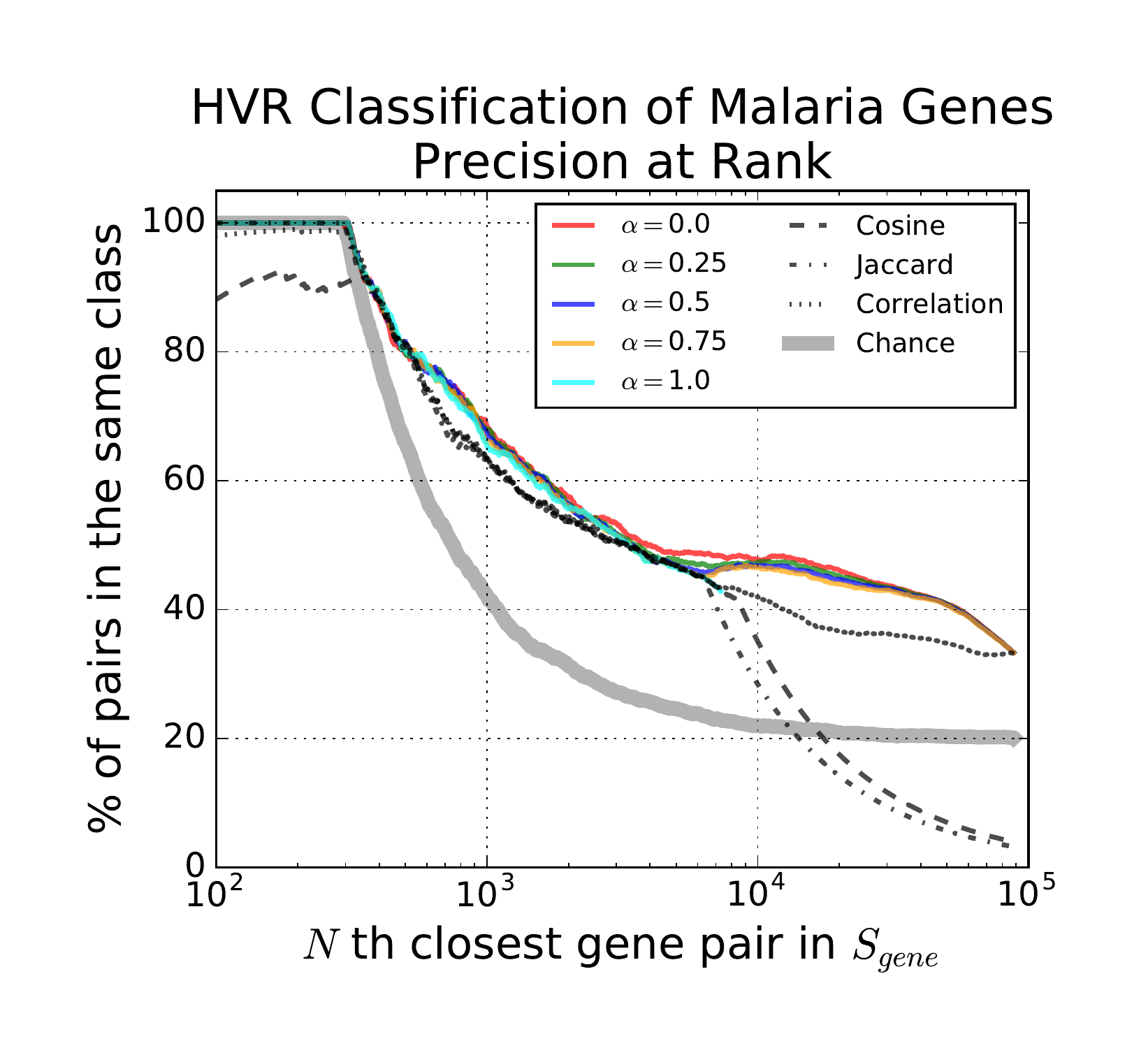}
  \caption{Precision at rank for the top $N$ most similar malaria genes in $S_{gene}$ (x axis; log-scale). Shown is the proportion (y axis) of the most similar gene pairs that are in the same HVR class, calculated using reflexive similarity ($L_{\infty}$-norm variant; colored lines), pairwise metrics (dotted black lines) and a null model that assigns genes to a random HVR class. Note that zero-valued elements in the similarity matrix (which varies across method) cause some method to fail at a certain rank (e.g. 7,000 for Jaccard similarity). The reflexive similarity measure, and some pairwise metrics, guarantee that genes are perfectly self-similar in the similarity matrix (the diagonal is entire 1-valued) making the first 297 scores 1.0. Beyond 297, the performance of each method deteriorates, and for some methods, drops below chance.}
  \label{fig:M_performance}
\end{figure}

Because there is only a ground-truth for the gene dimension, results are reported on $S_{gene}$. Figure \ref{fig:M_performance} shows the precision at rank: when pairs of genes are sorted in descending order by their similarity ($S_{i,j}$ for gene-pair $\langle i,j \rangle$), this is the proportion of pairs that are in the same HVR class. There are four classes in the gold standard and the null model simply assigns each gene to a random class, effectively making random-chance guesses. It should be noted that the adjacency matrix is not ordered in any systematic way, making this evaluation similar to the first task where $A$ was randomly permuted before similarity was computed. Second, while there was no discernible block-structure in $S_{gene}$ there may be a permutation that exposes such structure. This is, however, irrelevant: the task is to use the similarity scores generated on the gene-substring matrix to show that similar genes in $S_{gene}$ tend to be in the same HVR class.

The results show that reflexive variants with $\alpha < 1.0$ outperform pairwise metrics as well as the null model. Their performance is particularly better among less similar gene pairs, where some of the pairwise metrics assign perfect similarity to rows that have a single 1-valued element. This is evident in the drop-off observed with cosine and Jaccard measures near the 10,000th pair. By the final pair, the remaining methods converge to 45\%, which is the empirical number of gene-pairs that share an HVR class. When $\alpha=1.0$ and local structure has a strong effect, the results are significantly worse. This suggests that something about the substring dimension detracts from measuring gene-similarity. This may be to due to the nature of amino acids, where differences in the substrings are not uniformly metricated as character strings: a singular difference in one character may result in more distance than a difference somewhere else. This is further supported by the fact that there is only a slight degradation in performance as $\alpha$ approaches 1.0.

\begin{figure*}
  \centering
  \includegraphics[scale=0.375]{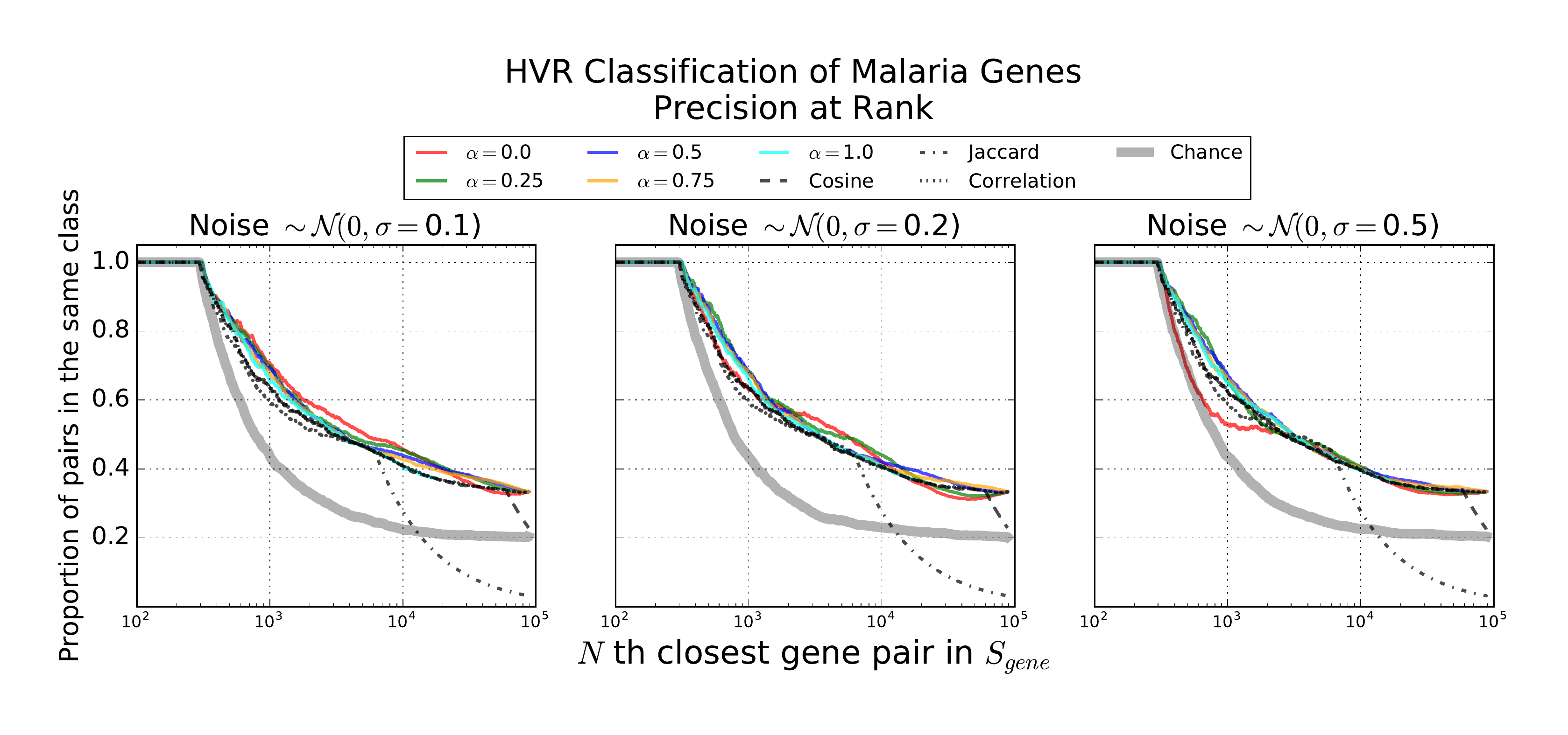}
    \caption{Precision at rank for versions of the malaria \textit{var} gene data with added noise. As the intensity of noise increases, all methods perform worse by rank 20,000. For low levels of noise, smaller values of $\alpha$ are helpful in the reflexive methods, but for noisier data, high values help more, particularly at lower ranks (more similar gene-pairs).}
  \label{fig:M_tilde_performance}
\end{figure*}

The malaria \textit{var} gene data were first made available in \cite{Thomas} and later analyzed in \cite{Larremore} which produced the HVR class assignments. We are not geneticists and are not able to assess the veracity of the original gene-substring data. Although it may be perfect, there is a good chance (perhaps due to experimental and measuring errors) that there is noise in the original adjacency structure. To simulate this with confidence, noise was added to the gene-substring matrix and the evaluation above was performed again. Figure \ref{fig:M_tilde_performance} shows results on data with noise of the form $\mathcal{N}(0,\sigma)$ for three values of $\sigma$. As expected, all methods perform worse than they did on the original data: whereas precision at rank 20,000 was about 49\% in the original data, with $\sigma=0.1$ it is about 40\%. This decline is exaggerated with greater intensity of added noise. With relatively low noise, the reflexive methods all out-perform the pairwise metrics, most prominently when $\alpha=0$. As the noise is increased, the performance of all methods diminishes, but it also becomes more similar. When $\sigma=0.5$, reflexive similarity with $\alpha \leq 0.5$, out-performs others for the top ranks (below 2,000), but by rank 20,000 only reflexive similarity for $\alpha=0.5$ and $\alpha=0.75$ are marginally better.

\section{Discussion}
Co-similar structure is found in many forms of bipartite data, but is sometimes taken for granted. Co-clustering analyses, like cluster analysis generally, often resort to a trial-and-error strategy when assessing latent clustered structure. The problem with such an analysis is two-fold: estimating the number of clusters is usually required before-hand, and the number of clusters in one dimension may be different than the other. Overall, our method of reflexive regular equivalence offers a way to attenuate the use of inter-dimensional structure and local transitivities in similarity calculations. Our results suggest that this is particularly important in noisy data. Reflexive similarity also offers a way to validate assumptions about co-similarity: if there is a permutation for $S$ that, when also applied to $S'$, exposes block structure, then $A$ is co-similar. In an experimental setting, the algorithm converges consistently with different values of $\alpha$ and for different adjacency structures. It remains unexplored the extent to which any properties of $A$ are required for the algorithm to convergence. Degenerate cases for $A$ may introduce numerical singularities during the reflexive algorithm. In such a case, pre-conditioning $A$ or the initializations of $S$ and $S'$ may be needed. But in addition to the rank of $A$, other structural properties may also bear on the performance of the algorithm.

\subsubsection*{Scalability}
The scalability of our method was only assessed experimentally here. This is primarily because the method's constituent representations and operations have well-understood scaling properties. In terms of representations, in many applications $A$ will be sparse, and if its sparsity distribution is similar in both dimensions, $S$ and $S'$ will also be sparse. All operations in Eqs. 9 and 10 can be performed sparsely \cite{Saad,Davis}, and have a number of high-performance, distributed implementations\footnote{ScaLAPACK; \url{www.netlib.org/scalapack}} \footnote{ARPACK; \url{www.caam.rice.edu/software/ARPACK}}. This allows the reflexive regular equivalence to take advantage of well-understood, salable foundation in linear algebra sub-routines. There is no matrix inversion which suggests that degeneracy in $A$ may not prevent convergence, though it may introduce instabilities in $S$ or $S'$. The convergence and runtime of the algorithm was assessed experimentally here and we leave a formal analysis to future work.

\subsubsection*{Generalization to $k$-partite structures}
Initial formulations of regular equivalence have been generalized to bipartite data \cite{Borgatti}. The formulation in Eqs. 9 and 10 can be generalized to $k$-partite structures without notational changes. In the case of $k$-partite structure, $A$ would be a rank-$k$ tensor. The reflexivity calculation, then, includes $k+1$-order item-wise product $A \otimes A^{\mathsf{T}}$. Though simple to notate, generalizing to $k$-partite structures requires two inner products between matrices (when $k=3$) or tensors ($k>3$) for each alternation of the algorithm.

The reflexive regular equivalence method described here offers an unsupervised way to explore potential co-similarity in noisy structures. The results on synthetic data and real-world malaria gene data show that our conception of reflexive equivalence is coherent (it actually measures similarity) and that it performs well compared to other metrics on asymmetric and noisy data. When faced with bipartite data of unknown structure, reflexive equivalence will help assess similarity in both dimensions. And by varying $\alpha$, it can enhance or reduce the contribution of the local equivalence component, essentially backing off to traditional spectral similarity in a two-mode / bipartite setting. In this way, our method offers a way to measure how co-similarity is helpful across dimensions, and thereby confirm or refute assumptions of co-similarity. Not only does reflexivity help overcome noise, by mixing regular equivalence and local structure, it helps pave the way to smarter exploratory analysis of bipartite data.

\section{Acknowledgments}
This project comprised part of a practicum for the Masters of Computer Science at the University of Chicago, lead by Borja Sotomayor and of which Zhou was a student. Shi and Gerow were supported by The Templeton Foundation grant to the Metaknowledge Research Network. Gerow is supported by the Trustees of the Isaak Walton Killam Foundation.

\bibliographystyle{acm}
\bibliography{clustering}  

\end{document}